\documentclass{article}

\usepackage{microtype}
\usepackage{graphicx}
\usepackage{subfigure}
\usepackage{booktabs} 

\usepackage{hyperref}

\usepackage[accepted]{icml2024}

\usepackage{amsmath}
\usepackage{amssymb}
\usepackage{mathtools}
\usepackage{amsthm}

\usepackage[capitalize,noabbrev]{cleveref}

\theoremstyle{plain}
\newtheorem{theorem}{Theorem}[section]

\newtheorem{corollary}[theorem]{Corollary}
\theoremstyle{definition}

\theoremstyle{remark}

\usepackage[textsize=tiny]{todonotes}

\icmltitlerunning{Physics in Next-token Prediction}

\begin{document}

\twocolumn[
\icmltitle{Physics in Next-token Prediction}



\icmlsetsymbol{equal}{\dag}
\icmlsetsymbol{correspondence}{*}

\begin{icmlauthorlist}
\icmlauthor{Hongjun An}{equal}
\icmlauthor{Yiliang Song}{equal}
\icmlauthor{Xuelong Li}{correspondence}
\end{icmlauthorlist}

\icmlcorrespondingauthor{Xuelong Li}{xuelong\_li@ieee.org}

\icmlkeywords{Physics, Next-token Prediction, Auto-regressive Model, Information Capacity}

\vskip 0.3in
]



\printAffiliationsAndNotice{\icmlEqualContribution} 

\begin{abstract}
We discovered the underlying physics in Next-token Prediction (NTP). We identified the law of information conservation within NTP and proposed the \textbf{First Law of Information Capacity (IC-1)}, demonstrating that the essence of intelligence emergence in auto-regressive models is fundamentally a process of information transfer. We also introduced Landauer's Principle into NTP, formulating the \textbf{Second Law of Information Capacity (IC-2)}, which establishes the relationship between auto-regressive model training and energy consumption. Additionally, we presented several corollaries, which hold practical significance for production practices. Finally, we demonstrate the consistency between the Law of Information Capacity and the Scaling Law for Neural Language Models, the Knowledge Capacity Scaling Laws, and the Scaling Laws for Precision.
\end{abstract}

\section{Introduction}

Currently, the state-of-the-art (SOTA) artificial intelligence models predominantly employ an auto-regressive architecture. Utilizing the Next-token Prediction (NTP) approach, these models seamlessly integrate various modalities, including text, images, audio, and video. This remarkable versatility and intelligence are reshaping human production and lifestyle in profound ways.

\begin{figure}[ht]
\vskip 0.2in
\begin{center}
\centerline{\includegraphics[width=0.93\columnwidth]{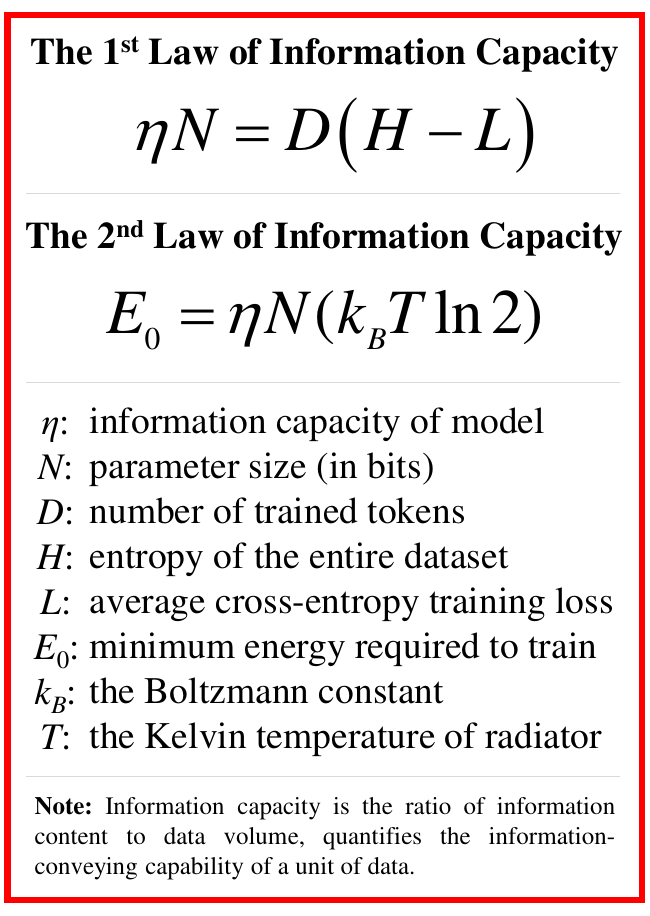}}
\caption{We propose the \textbf{First Law of Information Capacity (IC-1)} and the \textbf{Second Law of Information Capacity (IC-2)}, reveals the principle of information conservation and the energy relationship within NTP.}
\label{fig:eq}
\end{center}
\vskip -0.2in
\end{figure}

However, beneath this promising landscape looms a substantial ``\textit{scientific dark cloud.}'' Guided by the Scaling Law \cite{ScalingLaw1_2020}, researchers are relentlessly constructing ever-larger training datasets and expending increasing amounts of energy to train larger auto-regressive models in pursuit of more ``\textit{intelligent}'' systems. Why do big data and immense computational power lead to the emergence of ``\textit{intelligent}''? What is the essence behind this phenomenon? Where does the Scaling Law ultimately lead? What awaits us on the horizon?

In this paper, we delve deeply into these questions and uncover the underlying physics in NTP. We identify the law of information conservation \cite{hawking2014informationpreservationweatherforecasting} within NTP and derive the \textbf{First Law of Information Capacity (IC-1)}, demonstrating that the essence of intelligence emergence in auto-regressive models is fundamentally a process of information transfer. Additionally, we introduce Landauer's Principle \cite{5392446} into NTP, formulating the \textbf{Second Law of Information Capacity (IC-2)}, which reveals the relationship between the training process of auto-regressive models and the principles of energy in the real world. Based on our findings, we derive several corollaries that can effectively guide our daily practices and production activities. Finally, we validate the compatibility and complementarity of our insights with existing theories, attesting to the rationality of our theoretical framework.

The contribution of this paper can be summarized as follows:

\begin{itemize}
    \item We identified the conservation law of information in NTP and proposed the IC-1.
    \item We introduced the Landauer's Principle into NTP and proposed the IC-2.
    \item Based on IC-1 and IC-2, we derived several corollaries that can guide practical applications in human production.
    \item we demonstrate the consistency between our findings and the Scaling Law for Neural Language Models \cite{ScalingLaw1_2020}, the Knowledge Capacity Scaling Laws \cite{allenzhu2024physicslanguagemodels33}, and the Scaling Laws for Precision \cite{kumar2024scalinglawsprecision}.
\end{itemize}

\section{The First Law of Information Capability: the Conservation of Information in NTP}

\subsection{Preliminaries}

In 2023, Rae conceptualized the process of NTP as a mechanism for compressing information of dataset \cite{AGI_2023}, elucidating the relationship between the compression process and the emergence of intelligence: given a token vocabulary $\mathcal{V}$ and a dataset $\mathcal{D}=\{x_1, x_2, ..., x_{|\mathcal{D}|}\} (x_i \in \mathcal{V})$, our goal is to transmit $\mathcal{D}$ from $\mathcal{A}$ to $\mathcal{B}$ token-by-token as accurately as possible. During transmission, both parties can share an encoding and decoding function $f$. In this section, we will demonstrate that the intelligence level of function $f$ is related to its ability to compress $\mathcal{D}$.

\subsubsection{Baseline: Non-intelligent Transmission}

Assuming we have transmitted $\mathcal{D}_t=\{x_{1:t}\}(t<|\mathcal{D}|)$ and are about to transmit $x_{t+1}$, for a non-intelligent $f_0$, according to information theory \cite{Information_1948}, the length of the code $z_{t+1}^{f_0}=f_0(x_{t+1}|x_{1:t})$ is at least $-\log P(x_{t+1}|x_{1:t})$ (Eq. \ref{eq:len_zt}), which is its self-information.

\begin{equation}
\label{eq:len_zt}
    |z_{t+1}^{f_0}| = I(x_{t+1}|x_{1:t}) = -\log P(x_{t+1}|x_{1:t}),
\end{equation}

where the initial condition is $P(x_1|x_0)=P(x_1)$.

Thus, the total cost required by this method is as indicated by $I(\mathcal{D})$ (Eq. \ref{eq:cost_baseline}).

\begin{equation}
\label{eq:cost_baseline}
\begin{aligned}
    I(\mathcal{D}) &= \sum_{t=1}^{|\mathcal{D}|}{|z_t^{f_0}|} = \sum_{t=1}^{|\mathcal{D}|}-\log P(x_{t+1}|x_{1:t}) \\
    & =|\mathcal{D}|H(\mathcal{D}).
\end{aligned}
\end{equation}

\subsubsection{Intelligent Transmission based on Compression}

When auto-regressive models, such as large language models (LLMs), are applied to predict the next token, at each iteration, they input $x_{1:t}$ and predict $P(x_{t+1}|x_{1:t}, f_a)$. So the total cost required by this method to transmit $\mathcal{D}$ is at least $I(\mathcal{D}|f_a)$ (Eq. \ref{eq:cost_auto}).

\begin{equation}
\label{eq:cost_auto}
\begin{aligned}
    & I(\mathcal{D}|f_a) = \sum_{t=0}^{|\mathcal{D}|-1} -\log P(x_{t+1}|x_{1:t}, f_a).
\end{aligned}
\end{equation}

It is noteworthy that the Eq. \ref{eq:cost_auto} is in exact correspondence with the objective loss function, cross-entropy loss, which is optimised during the training phase (Eq. \ref{eq:loss_lms}).

\begin{equation}
\label{eq:loss_lms}
\begin{aligned}
    \ell(f_a) & = \frac{1}{|\mathcal{D}|}\sum_{t=0}^{|\mathcal{D}|-1} -\log P(x_{t+1}|x_{1:t}, f_a) \\
    & = \frac{1}{|\mathcal{D}|} I(\mathcal{D}|f_a).
\end{aligned}
\end{equation}

Thus, the model training process, characterized by the reduction of the loss function, can be regarded as a compression process of the dataset $\mathcal{D}$. The higher the compression, the more intelligent of the model becomes.

\subsection{A Derivation of the First Law of Information Capability}\label{subsec:derivation_lse}

If we conduct a meticulous examination of Eq. \ref{eq:cost_baseline} and Eq. \ref{eq:loss_lms}, we may uncover an even more intriguing insight. That is, when $f_a$ is sufficiently powerful, $I(\mathcal{D}|f_a)$ is absolutely likely to be smaller than $I(\mathcal{D})$. Does this imply that some information disappears into thin air (Eq. \ref{eq:dis_info})?

\begin{equation}
\label{eq:dis_info}
     I(\mathcal{D}) - I(\mathcal{D}|f_a) = \text{?}
\end{equation}

In fact, from the perspective of physics, \textbf{information is conserved} \cite{hawking2014informationpreservationweatherforecasting}, these pieces of information do not vanish; rather, they are transferred into the model $f_a$ (Eq. \ref{eq:info_conservation}). 

\begin{equation}
\label{eq:info_conservation}
\begin{aligned}
    I(f_a^{+}) = I(\mathcal{D}) - I(\mathcal{D}|f_a),
\end{aligned}
\end{equation}

where $I(f_a^{+})$ represents the effective information stored in $f_a$ pertaining to task $\mathcal{D}$. We introduce the concept of \textbf{Information Capacity} \cite{distillation_2021}, denoted by $\eta$, defined as the ratio of effective information $I(f_a^{+})$ to its data size $N$ (approximately equal to the number of parameter size, measured in bits). Thus, we obtain Eq. \ref{eq:scaling_law_pre}

\begin{equation}
\label{eq:scaling_law_pre}
\begin{aligned}
    \eta = \frac{I(f_a^{+})}{N}.
\end{aligned}
\end{equation}

After performing equivalent transformations to the Eq. \ref{eq:scaling_law_pre}, we arrive at the \textbf{First Law of Information Capacity (IC-1)}, as shown in Eq. \ref{eq:scaling_law}.

\begin{equation}
\label{eq:scaling_law}
\begin{aligned}
    \eta N = D(H-L).
\end{aligned}
\end{equation}

where $L=\ell(f_a)$ denotes the average cross-entropy loss, $D=|\mathcal{D}|$ denotes the number of tokens in dataset $\mathcal{D}$, and $H=H(\mathcal{D})$ denotes the entropy of the entire dataset.

Therefore, the process of model training is essentially a process of compressing dataset $\mathcal{D}$, leading to a reduction in the loss function, the transfer of information to model $f_a$, and an increase in its information capacity $\eta$. This transfer is driven by the model training process, particularly the back propagation algorithm \cite{BP_1986}. The energy for this transfer comes from electrical resources in the real world (Sec. \ref{sec:law_of_ce}).

\subsection{A Dynamic Perspective to the Process of Intelligence Emergence}\label{subsec:dynamic_intelligence_emergence}

Currently, Eq. \ref{eq:scaling_law} remains \textbf{static}. In particular, within our derivation, $D$ represents the total number of tokens in the dataset, $H$ denotes the overall entropy of the dataset, and $L$ is the overall average loss after training has concluded. From the perspective of the law of conservation of information, it is imperative that conservation be observed not only at the terminal state, but throughout the entirety of the \textbf{dynamic} training process. Therefore, we will restate the meanings of the variables in Eq. \ref{eq:scaling_law}:

\begin{itemize}
    \item $H$: the overall entropy of the dataset, is a constant.
    \item $N$: parameter size of the model, measured in bits. Once the model architecture is established, it will become a fixed constant.
    \item $D$: the token numbers that \textbf{has been trained}. This is a \textbf{variable} that monotonically increases as training progresses.
    \item $\eta,L$: $\eta$ is the information capacity of the model, and $L$ is the dynamic average cross-entropy loss. Both change dynamically as $D$ increases.
\end{itemize}

It is therefore possible to describe the entirety of the model training process from a dynamic perspective based on IC-1:

\paragraph{Initial State.} Training has not yet begun, with no information transfer, thus $\eta=0$. 

\paragraph{Training State.} As training progresses, $D$ gradually increases, leading to a decrease in $L$. To satisfy the equation, $\eta$ must inevitably increase. During this dynamic process, the information gradually transfers to the $f_a$, prompting the model to \textit{learn}.

\paragraph{Terminal State.} When $\eta=\eta_{\text{max}}$ (determined by the model architecture), the information that the model parameters can store reaches saturation. At this point, continuing the training process does not enable the model to learn more tokens; $L$ converges, and training concludes.

\section{The Second Law of Information Capacity: the Energy Relationship in NTP}\label{sec:law_of_ce}

In Sec. \ref{subsec:derivation_lse}, the IC-1 indicates that the learning process in auto-regressive models is fundamentally an information transfer process. The driving force behind this transfer comes from the back-propagation \cite{BP_1986} algorithm, while the energy is sourced from the power of the physical world. One might wonder, what is the minimum amount of energy required to complete this information transfer process?

In 1961, Landauer proposed that the energy required to erase a single bit is at least 
$k_B T \ln 2$, known as the \textbf{Landauer's Principle} \cite{5392446}. Therefore, according to Eq. \ref{eq:scaling_law}, when we transfer information $I(f_a^{+})$, at least energy $E_0=I(f_a^{+}) k_B T \ln 2$ must be consumed. Thus, the training process of auto-regressive models establishes an energy relationship with the physical world. We can derive the \textbf{Second Law of Information Capacity (IC-2)} (Eq. \ref{eq:landauer}).

\begin{equation}
\label{eq:landauer}
    E_0 = \eta N (k_B T \ln 2),
\end{equation}

where $E_0$ is the minimum energy required to complete the information transfer, $k_B \approx 1.38 \times 10^{-23} \text{J/K}$ is the Boltzmann constant, and $T$ is the temperature of the heat reservoir in Kelvin.

\section{Corollaries to Guide Practice}

\subsection{Estimating the Entropy of the Dataset}

From Sec. \ref{subsec:dynamic_intelligence_emergence}, it can be inferred that when the model training process is in its \textbf{initial state}, $\eta \approx 0$. From this, we can derive the following corollary: the Data Entropy Estimation Theorem (Corollary \ref{coro:data_entropy_estimation}).

\begin{corollary}
\label{coro:data_entropy_estimation}
    The entropy of the dataset is approximately equal to the initial loss of the model training.

    Proof.

    \begin{equation}
    \begin{aligned}
        & \eta N = D(H-L), D>0, \eta \approx 0, \\
        & \Rightarrow H \approx L.
    \end{aligned}
    \end{equation}
\end{corollary}

It is crucial to acknowledge that this corollary is exclusively applicable for estimation purposes. This is due to the fact that the model itself incorporates initial information following its initialisation, which ultimately leads to the emergence of $\eta>0$.

\subsection{Estimating the Quality of the Dataset}

When the number of tokens in the dataset is fixed, a higher entropy of the dataset indicates that it can provide more information for the model to learn. Therefore, we can consider entropy $H$ as a metric for evaluating dataset quality, with the assessment method outlined in Corollary \ref{coro:data_entropy_estimation}.

\subsection{Matching Suitable Model Size with Dataset Size}\label{subsec:match}

In the practice of large model production, determining the appropriate model size, the amount of training data required, and the duration of training is a critical issue. In previous work, the Knowledge Capacity Scaling Laws indicated that current large language models (LLMs) generally store only 2 bits of information per parameter \cite{allenzhu2024physicslanguagemodels33}. For models using the FP16/BF16 or INT8 formats to store parameters, the information capacity $\eta$ is approximately $0.125 \sim 0.25$. Additionally, according to the technical reports of well-known large models \cite{ScalingLaw1_2020}, it has been inferred that the entropy of datasets typically ranges from 10 to 15 approximately. Thus, if $N$ and $L$ are given, the required dataset size $D$ can be estimated; if $D$ and $L$ are given, the required model size can be estimated.

Sec. \ref{subsec:consis_scaling_law} indicates that the IC-1 is compatible with the Scaling Law proposed by the OpenAI team in 2020 \cite{ScalingLaw1_2020}. Their advantage lies in providing power-law relationships between $L$ and $N$, $L$ and $D$, as well as $L$ and $C$ (computation cost) respectively through extensive experiments. However, these power-law relationships are validated only for large models based on the Transformer architecture. From the IC-1, we cannot derive the relationships between $L$ and $N$ or $D$ respectively; we can only calculate their corresponding values from a macro perspective. Nevertheless, the IC-1 is applicable to all auto-regressive models, regardless of architecture, because its underlying principle is based on the physical principle of information conservation. Therefore, we are compatible and complementary with the OpenAI's Scaling Law.

\subsection{Identify the Energy Limits Required for Training Auto-regressive Models}

As chip technology advances and autore-gressive model learning algorithms are optimized, the energy required to train models achieving similar intelligence levels is expected to decline. Additionally, future developments may lead to a shift from GPUs to quantum computers, which could further reduce energy consumption. Nevertheless, regardless of technological advancement, there is an inherent limitation to the amount of energy that can be consumed for the transmission of information, known as Landauer's Limit, which is represented by IC-2 (Eq. \ref{eq:landauer}).

\subsection{The conditions for lossless quantization}

Quantizing the weights of autoregressive models to achieve economic goals is a common practice. However, achieving lossless quantization requires certain conditions. Suppose the model weights before quantization are $b$-bit with information capacity $\eta$,  and after quantization, they are $b'$-bit with information capacity $\eta '$. To achieve lossless quantization, the following conditions must be met Eq. \ref{eq:con_qua}:

\begin{equation}
\label{eq:con_qua}
    \eta b \leq \eta' b'.
\end{equation}

Since $\eta' \leq 1$, we can derive the necessary condition for lossless quantization as Eq. \ref{eq:con_qua_co}:

\begin{equation}
\label{eq:con_qua_co}
    \eta b \leq b'.
\end{equation}

\section{Consistency with Existing Theories}

\subsection{Consistency with the Scaling Law of Neural Language Models}\label{subsec:consis_scaling_law}

In this section, we will substitute the experimental data from \cite{ScalingLaw1_2020} into the IC-1 to verify the consistency between the IC-1 and the Scaling Law of Neural Language Models. According to Sec. \ref{subsec:dynamic_intelligence_emergence}, we can know that the $H$ is a constant.

\paragraph{When $L$ is fixed, }it indicates that model training has stopped. At this point, for a given set of $N$ and $D$—that is, for a specific model trained on a defined number of tokens—$\eta$ becomes a constant. Consequently, $(H-L)/\eta$ is also a constant, resulting in $N$ being proportional to $D$.

\begin{equation}
\begin{aligned}
    & N=kD,k=\frac{H-L}{\eta}; \\
    & \Rightarrow N \propto D.
\end{aligned}
\end{equation}

\paragraph{When $D$ is fixed, } it implies that the same number of tokens is being trained. If the model structure is altered to increase $N$, the change in the information capacity $\eta$ cannot be directly determined. Consequently, the relationship between $L$ and $N$ cannot be directly established (Eq. \ref{eq:l_relation}).

\begin{figure*}[t]
\vskip 0.2in
\begin{center}
\centerline{\includegraphics[width=1.6\columnwidth,clip,trim={20pt 20pt 20pt 30pt}]{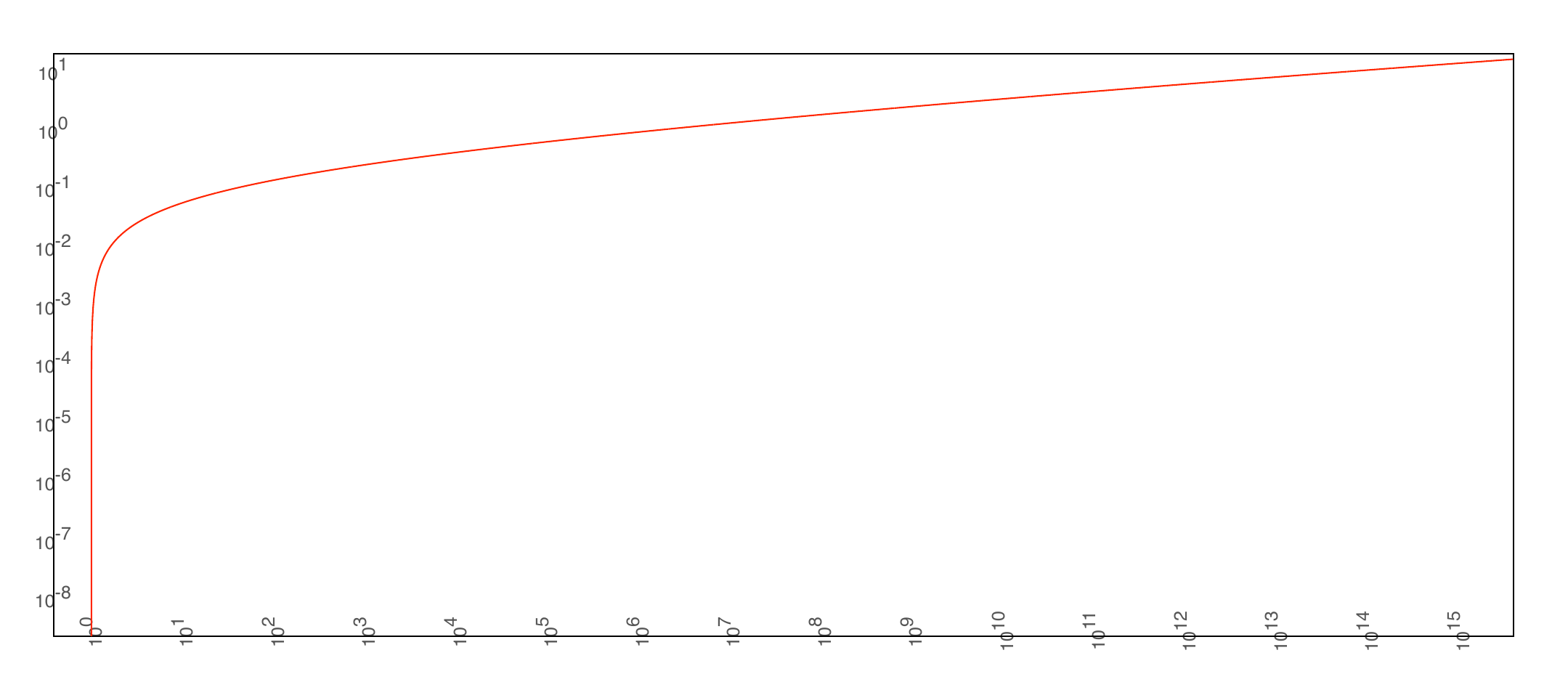}}
\caption{The plotted curve of $f(x)=x^{0.095}-x^{0.076}$. When $0 < x \ll 10^{15} $, $0 < f(x) \ll 10$.}
\label{fig:scale_func}
\end{center}
\vskip -0.2in
\end{figure*}

\paragraph{When $N$ is fixed, }it implies that the model is deterministic. If $D$ is increased, meaning that the model is trained on a larger number of tokens, $\eta$ will also increase correspondingly. However, the relationship between the change rate of the two cannot be directly determined, and thus the relationship between $L$ and $D$ also cannot be established directly (Eq. \ref{eq:l_relation}).

\begin{equation}
\label{eq:l_relation}
    L = H-\eta \frac{N}{D}.
\end{equation}

\paragraph{In summary, }to verify whether the IC-1 aligns with the empirical formula measured by the OpenAI team in 2020, we can approach the analysis from the perspective of fixed $L$. Here is the empirical formula (Eq. \ref{eq:relation_L_D} and Eq. \ref{eq:relation_L_N}):

\begin{equation}
\label{eq:relation_L_D}
\begin{aligned}
    & L = \left(\frac{D}{5.4 \times 10^{13}}\right)^{-0.095}, 
\end{aligned}
\end{equation}

\begin{equation}
\label{eq:relation_L_N}
\begin{aligned}
    & L = \left(\frac{\frac{1}{16}N}{8.8 \times 10^{13}}\right)^{-0.076}. 
\end{aligned}
\end{equation}

The factor of $\frac{1}{16}$ in the Eq. \ref{eq:relation_L_N} arises because the parameter count in the original paper refers to the quantity of FP/BF16 numbers, whereas $N$ in this paper is expressed in bits. By setting (\ref{eq:relation_L_D})$=$(\ref{eq:relation_L_N}), we obtain:

\begin{equation*}
\begin{aligned}
    \left(\frac{5.4 \times 10^{13}}{D}\right)^{0.095} = \left(\frac{8.8 \times 10^{13}}{\frac{1}{16}N}\right)^{0.076}.
\end{aligned}
\end{equation*}

By constructing the function $f(x)=x^{0.095}-x^{0.076}$ and plotting its curve (Fig. \ref{fig:scale_func}), it becomes evident that when $0 < x \ll 10^{15} $, $0 < f(x) \ll 10$. Therefore, it can be inferred that the bases on both sides of the equation are nearly equal, with the difference in exponents attributed to observational errors.

\begin{equation*}
\begin{aligned}
    \frac{5.4 \times 10^{13}}{D} \approx \frac{8.8 \times 10^{13}}{\frac{1}{16}N}.
\end{aligned}
\end{equation*}

Thus, we can deduce that:

\begin{equation*}
\begin{aligned}
    N \approx 26.08D.
\end{aligned}
\end{equation*}

This is consistent with the IC-1. Furthermore, when $\eta = 0$, meaning the model is untrained, we have $L_0 = H \approx 10$ (as visually inferred from Figure 2 in the paper \cite{ScalingLaw1_2020}); when the model converges, $L \in [2,7] $. Thus, we obtain:

\begin{equation*}
\begin{aligned}
    \eta = \frac{H-L}{k} \in \left[\frac{10-7}{26.08}, \frac{10-3}{26.08} \right] = [0.115, 0.268].
\end{aligned}
\end{equation*}

The information capacity $\eta$ falls within a normal range ($<1$), thus indicating that the IC-1 is consistent with OpenAI's empirical formula.

\subsection{Consistency with the Knowledge Capacity Scaling Laws}

As noted in Sec. \ref{subsec:match}, the information capacity of LLMs in the Knowledge Capacity Scaling Laws should generally be less than $0.125\sim 0.25$ \cite{allenzhu2024physicslanguagemodels33}, which is consistent with the value of $[0.115,0.268]$ in Sec. \ref{subsec:consis_scaling_law}.

\subsection{Consistency with the Scaling Laws for Precision}

\cite{kumar2024scalinglawsprecision} found that as the model is trained on more data, the degradation caused by post-training quantization increases, ultimately making additional pre-training data detrimental. This is, in fact, an inevitable consequence of the First Law of Information Capacity. As the model is trained on more data, the information capacity $\eta$ of the model will gradually increase. When condition \ref{eq:con_qua_co} is violated, quantization of the model will damage the information learned by the model, leading to negative effects.

\section{Conclusion}

This paper has revealed the fundamental physical principles that underpin NTP by establishing the IC-1 and the IC-2. These laws elucidate not only the conservation of information and the energy requirements in auto-regressive model training, but also offer practical corollaries that can guide the development and training of intelligent systems. By aligning our findings with existing theories, we have demonstrated the broad applicability and significance of our theoretical framework, thereby paving the way for more efficient and sustainable advancements in artificial intelligence.


\bibliography{agi}
\bibliographystyle{icml2024}


\end{document}